\documentclass[10pt,twocolumn,letterpaper]{article}

\usepackage{cvpr}
\usepackage{times}
\usepackage{epsfig}
\usepackage{graphicx}
\usepackage{amsmath}
\usepackage{amssymb}
\usepackage{mathtools}

\usepackage{adjustbox}
\usepackage{bm,upgreek}
\usepackage{booktabs}
\usepackage{graphicx,multirow}
\usepackage{balance}

\usepackage{url}
\usepackage{multirow}
\usepackage{array}

\usepackage[pagebackref=true,breaklinks=true,letterpaper=true,colorlinks,bookmarks=false]{hyperref}

\cvprfinalcopy 

\def\cvprPaperID{312} 

\ifcvprfinal\fi
\begin{document}

\title{Diversity Regularized Spatiotemporal Attention\\for Video-based Person Re-identification}



\newcommand*\samethanks[1][\value{}]{\footnotemark[#1]}
\author{
Shuang Li$^{1}$\thanks{Work done while employed at Disney Research Pitsburgh.}\and Slawomir Bak$^{2}$\samethanks[1] \and Peter Carr$^{2}$\samethanks[1] \and Xiaogang Wang$^{1}$ \and
\begin{tabular}{*{2}{>{\centering}p{.4\textwidth}}}
\small  &  \tabularnewline
$^{1}$CUHK-SenseTime Joint Lab,\\ The Chinese University of Hong Kong & $^{2}$Argo AI \tabularnewline
{\tt\small \{sli,xgwang\}@ee.cuhk.edu.hk} & {\tt\small \{sbak,pcarr\}@argo.ai}
\end{tabular}
}

\maketitle
\thispagestyle{empty}

\begin{abstract}
Video-based person re-identification matches video clips of people across non-overlapping cameras. Most existing methods tackle this problem by encoding each video frame in its entirety and computing an aggregate representation across all frames. In practice, people are often partially occluded, which can corrupt the extracted features. Instead, we propose a new spatiotemporal attention model that automatically discovers a diverse set of distinctive body parts.  This allows useful information to be extracted from all frames without succumbing to occlusions and misalignments.  The network learns multiple spatial attention models and employs a diversity regularization term to ensure multiple models do not discover the same body part.  Features extracted from local image regions are organized by spatial attention model and are combined using temporal attention. As a result, the network learns latent representations of the face, torso and other body parts using the best available image patches from the entire video sequence. Extensive evaluations on three datasets show that our framework outperforms the state-of-the-art approaches by large margins on multiple metrics.
\end{abstract}

\begin{figure}[t]
\begin{center}
\hspace{-12pt}
\includegraphics[width=1.05\linewidth]{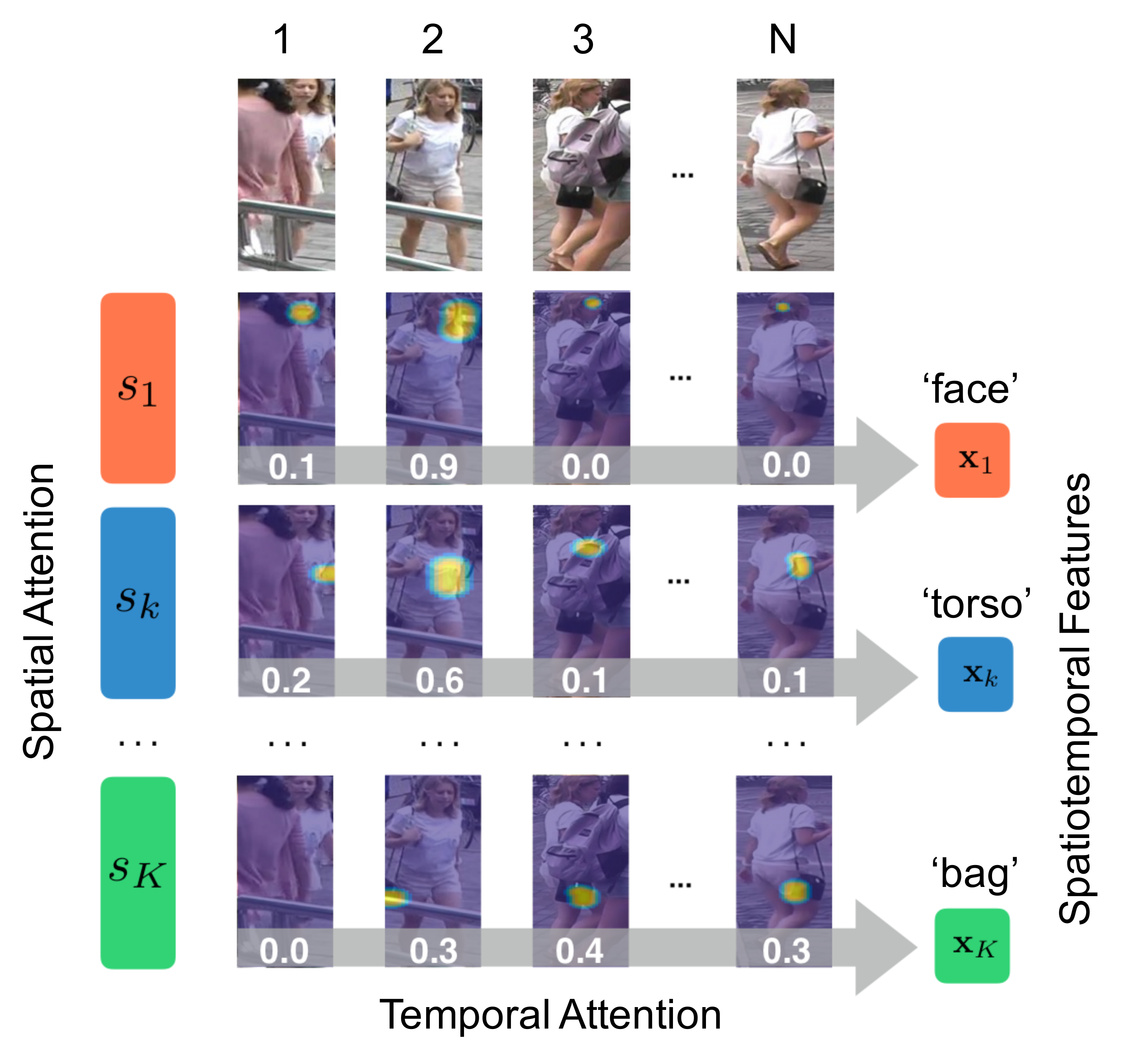} \ \\
\end{center}
\vspace{-10pt}
\caption{
\textbf{Spatiotemporal Attention.} In challenging video re-identification scenarios, a person is rarely fully visible in all frames.  However, frames in which only part of the person is visible often contain useful information.  For example, the face is clearly visible in the frames 1 and 2, the torso in frame 2, and the handbag in frames 2, 3 and $N$.  Instead of averaging full frame features across time, we propose a new spatiotemporal approach which learns to detect a set of $K$ diverse salient image regions within each frame (superimposed heatmaps).  An aggregate representation of each body part is then produced by combining the extracted per-frame regions across time (weights shown as white text). Our spatiotemporal approach creates a compact encoding of the video that exploits useful partial information in each frame by leveraging multiple spatial attention models, and combining their outputs using multiple temporal attention models.}
\label{intro}
\end{figure}

\section{Introduction}
\label{sec:intro}
Person re-identification matches images of pedestrians in one camera with images of pedestrians from another, non-overlapping camera. This task has drawn increasing attention in recent years due to its importance in applications, such as  surveillance \cite{wang2013intelligent}, activity analysis \cite{loy2009multi} and tracking \cite{yu2013harry}. It remains a challenging problem because of complex variations in camera viewpoints, human poses, lighting, occlusions, and background clutter.

In this paper, we investigate the problem of video-based person re-identification, which is a generalization of the standard image-based re-identification task.  Instead of matching image pairs, the algorithm must match pairs of video sequences (possibly of different durations).  A key challenge in this paradigm is developing a good latent feature representation of each video sequence.

Existing video-based person re-identification methods represent each frame as a feature vector and then compute an aggregate representation across time using average or maximum pooling \cite{zhousee,liu2017video,you2016top}.  Unfortunately, this approach has several drawbacks when applied to datasets where occlusions are frequent (Fig.~\ref{intro}).  The feature representation generated for each image is often corrupted by the visual appearances of occluders.  However, the remaining visible portions of the person may provide strong cues for re-identification.  Assembling an effective representation of a person from these various glimpses should be possible.  However, aggregating features across time is not straightforward.  A person's pose will change over time, which means any aggregation method must account for spatial misalignment (in addition to occlusion) when comparing features extracted from different frames.

In this paper, we propose a new spatiotemporal attention scheme that effectively handles the difficulties of video-based person re-identification.  Instead of directly encoding the whole image (or a predefined decomposition, such as a grid), we use multiple spatial attention models to localize discriminative image regions, and pool these extracted local features across time using temporal attention.  Our approach has several useful properties:
\begin{itemize}
    \item Spatial attention explicitly solves the alignment problem between images, and avoids features from being corrupted by occluded regions. 
    \item Although many discriminative image regions correspond to body parts, accessories like sunglasses, backpacks and hats; are prevalent and useful for re-identification.  Because these categories are hard to predefine, we employ an unsupervised learning approach and let the neural network automatically discover a set of discriminative object part detectors (spatial attention models). 
    \item We employ a novel diversity regularization term based on the Hellinger distance to ensure multiple spatial attention models do not discover the same body part.
    \item We use temporal attention models to compute an aggregate representation of the features extracted by each spatial attention model. These aggregate representations are then concatenated into a final feature vector that represents all of the information available from the entire video.
\end{itemize}

We demonstrate the effectiveness of our approach on three challenging video re-identification datasets.  Our technique out performs the state-of-the-art methods under multiple evaluation metrics.


\section{Related Work}

Person re-identification was first proposed for multi-camera tracking \cite{wang2013intelligent,shen2014multihuman}. Gheissari \etal \cite{gheissari2006person} designed a spatial-temporal segmentation method to extract visual cues and employed color and salient edges for foreground detection. This work defined the image-based person re-identification as a specific computer vision task. 

\textbf{Image-based person re-identification}
mainly focuses on two categories: extracting discriminative features \cite{gray2008viewpoint,farenzena2010person,ma2012local,kviatkovsky2013color,xiaoend} and learning robust metrics \cite{prosser2010person,zheng2011person,koestinger2012large,pedagadi2013local,bakone}. In recent years, researchers have proposed numerous deep learning based methods \cite{ahmed2015improved,li2014deepreid,ding2015deep,li2017learning,xiao2017joint} to jointly handle both aspects. Ahmed \etal \cite{ahmed2015improved} input a pair of cropped pedestrian images to a specifically designed CNN with a binary verification loss function for person re-identification. In \cite{ding2015deep}, Ding \etal minimize feature distances between the same person and maximize the distances among different people by employing a triplet loss function when training deep neural networks. Xiao \etal \cite{xiao2017joint} jointly train the pedestrian detection and person re-identification in a single CNN model. They propose an Online Instance Matching loss function which learns features more efficiently in large scale verification problems.

\textbf{Video-based person re-identification.}
Video-based person re-identification \cite{mclaughlin2016recurrent,zhousee,you2016top,wang2016person,zhu2016video,ma2017person} is an extension of  image-based approaches.  Instead of pairs of images, the learning algorithm is given pairs of video sequences. In \cite{you2016top}, You \etal present a top-push distance learning model accompanied by the minimization of intra-class variations to optimize the matching accuracy at the top rank for person re-identification. McLaughlin \etal \cite{mclaughlin2016recurrent} introduce an RNN model to encode temporal information. They utilize temporal pooling to select the maximum activation over each feature dimension and compute the feature similarity of two videos. Wang \etal \cite{wang2016person} select reliable space-time features from noisy/incomplete image sequences while simultaneously learning a video ranking function. Ma \etal \cite{ma2017person} encode multiple granularities of spatiotemporal dynamics to generate latent representations for each person. A Time Shift Dynamic Time Warping model is derived to select and match data between inaccurate and incomplete sequences.

\textbf{Attention models for person re-identification.}
Attention models \cite{xu2015show, Li_2017_CVPR, Li_2017_ICCV} have grown in popularity since \cite{xu2015show}. Zhou \etal \cite{zhousee} combine spatial and temporal information by building an end-to-end deep neural network. An attention model assigns importance scores to input frames according to the hidden states of an RNN. The final feature is a temporal average pooling of the RNN's outputs.
However, if trained in this way, corresponding weights at different time steps of the attention model tend to have the same values. 
Liu \etal \cite{liu2017hydraplus} proposed a multi-directional attention module to exploit the global and local contents for image-based person re-identification. However, jointly training multiple attentions might cause the mode collapse. The network has to be carefully trained to avoid attention models focusing on similar regions with high redundancy. 
In this paper, we combine spatial and temporal attentions into spatiotemporal attention models to address the challenges in video-based person re-identification. 
For spatial attention, we use a penalization term to regularize multiple redundant attentions. 
We employ temporal attention to assign weights to different salient regions on a per-frame basis to take full advantage of discriminative image regions. Our method demonstrates better empirical performance, and decomposes into an intuitive network architecture.

\section{Method}

\begin{figure*}[t]
\begin{center}
\includegraphics[width=1.0\linewidth]{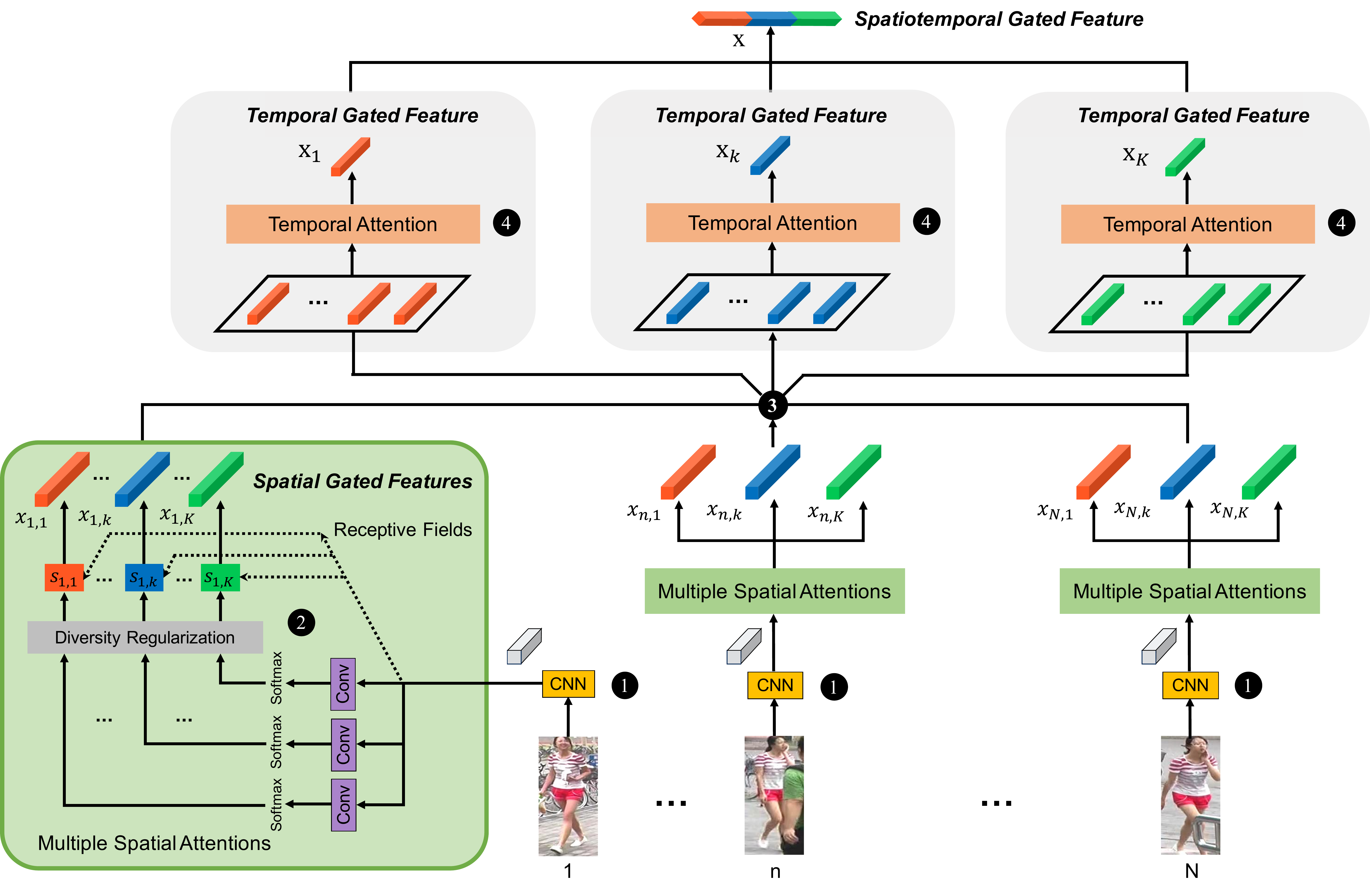} \ \\
\end{center}
\vspace{-3mm}
\caption{\textbf{Spatiotemporal Attention Network Architecture}. The input video is reduced to $N$ frames using restricted random sampling. (1) Each image is transformed into feature maps using a CNN. (2) These feature maps are sent to a conventional network followed by a softmax function to generate multiple spatial attention models and corresponding receptive fields for each input image.  A diversity regularization term encourages learning spatial attention models that do not result in overlapping receptive fields per image. Each spatial attention model discovers a specific salient image region and generates a spatial gated feature (Fig.~\ref{vis}). (3) Spatial gated features from all frames are grouped by spatial attention model. (4) Temporal attentions compute an aggregate representation for the set of features generated by each spatial attention model. Finally, the spatiotemporal gated features for all body parts are concatenated into a single feature which represents the information contained in the entire video sequence. }
\label{net2}
\end{figure*}

We propose a new deep learning architecture (Fig.~\ref{net2}) to better handle video re-identification by automatically organizing the data into sets of consistent salient subregions. Given an input video sequence, we first use a restricted random sampling strategy to select a subset of video frames (Sec.~\ref{sample}). Then we send the selected frames to a multi-region spatial attention module (Sec.~\ref{sec:mrsa}) to generate a diverse set of discriminative spatial gated visual features---each roughly corresponding to a specific salient region of a person (Sec.~\ref{sec:diversity}). 
The overall representation of each salient region across the duration of the video is generated using temporal attention (Sec.~\ref{sec:temporal}). Finally, we concatenate all temporal gated features and send them to a fully-connected layer which represents the latent spatiotemporal encoding of the original input video sequence. An OIM loss function, proposed by Xiao \etal \cite{xiao2017joint}, is built on top of the FC layer to supervise the training of the whole network in an end-to-end fashion. However, any traditional loss function (like softmax) could also be employed.

\subsection{Restricted Random Sampling}
\label{sample}

Previous video-based person re-identification methods \cite{mclaughlin2016recurrent,ma2017person,zhousee} do not model long-range temporal structure because the input video sequences are relatively short.  To some degree, this paradigm is only slightly more complicated than image-based re-identification since consecutive video frames are highly correlated, and the visual features extracted from one frame do not change drastically over the course of a short sequence. However, when input video sequences are long, any re-identification methodology must be able to cope with significant visual changes over time, such as different body poses and angles relative to the camera.  

Wang \etal \cite{wang2016temporal} proposed a temporal segment network to generate video snippets for action recognition. 
Inspired by them, we propose a restricted random sampling strategy to generate compact representations of long video sequences that still provide good representations of the original data. Our approach enables models to utilize visual information from the entire video and avoids the redundancy between sequential frames.
Given an input video $\mathbf{V}$, we divide it into $N$ chunks $\{C_n\}_{n=1,N}$ of equal duration. From each chunk $C_n$, we randomly sample an image $I_n$. The video is then represented by the ordered set of sampled frames $\{ I_n \}_{n=1,N}$. 


\subsection{Multiple Spatial Attention Models}
\label{sec:mrsa}
We employ multiple spatial attention models to automatically discover salient image regions (body parts or accessories) useful for re-identification.
Instead of pre-defining a rigid spatial decomposition of input images (\eg a grid structure), our approach automatically identifies multiple disjoint salient regions in each image that consistently occur across multiple training videos.  Because the network learns to identify and localize these regions (e.g. automatically discovering a set of object part detectors), our approach mitigates registration problems that arise from pose changes, variations in scale, and occlusion. 
Our approach is not limited to detecting human body parts. It can focus on any informative image regions, such as hats, bags and other accessories often found in re-identification datasets. 
Feature representations directly generated from entire images can easily miss fine-grained visual cues (Fig.~\ref{intro}).  
Multiple diverse spatial attention models, on the other hand, can simultaneously discover discriminative visual features while reducing the distraction of background contents and occlusions. Although spatial attention is not a new concept, to the best of our knowledge, this is first time that a network has been designed to automatically discover a diverse set of attentions within image frames that are consistent across multiple videos.

As shown in Fig.~\ref{net2}, we adopt the ResNet-50 CNN architecture \cite{he2016deep} as our base model for extracting features from each sampled image. The CNN has a convolutional layer in front (named $conv1$), followed by four residual blocks. We exploit $conv1$ to $res5c$ as the feature extractor.  As a result, each image $I_n$ is represented by an $8\times4$ grid of feature vectors $\{ \mathbf{f}_{n,\ell} \}_{\ell=1,L}$, where $L=32$ is the number of grid cells, and each feature is a $D=2048$ dimensional vector.


Multiple attention models are then trained to locate discriminative image regions (distinctive object parts) within the training data. For the $k^\text{th}$ model, $k \in (1, \ldots, K)$, the amount of spatial attention $s_{n,k,\ell}$ given to the feature vector in cell $\ell$ is based on a response $e_{n,k,\ell}$ generated by passing the feature vector through two linear transforms and a ReLU activation in between. Specifically, 
\begin{align}
e_{n,k,\ell} = (\mathbf{w}_{s,k}')^\mathsf{T} \max( \mathbf{W}_{s,k} \mathbf{f}_{n,\ell} + \mathbf{b}_{s,k}, 0) + b_{s,k}',
\label{eqn:spatial-activation}
\end{align}

\noindent where $\mathbf{w}'_{s,k} \in \mathbb{R}^{d}$, $\mathbf{W}_{s,k} \in \mathbb{R}^{d \times D}$, $\mathbf{b}_{s,k} \in \mathbb{R}^{d}$ and $b_{s,k}' \in \mathbb{R}$ are parameters to be learned for the $k^\text{th}$ spatial attention model. The first linear transform projects the original feature to a lower $d=256$ dimensional space, and the second transform produces a scalar value for each feature/cell. The attention for each feature/cell is then computed as the softmax of the responses
\begin{align}
s_{n,k,\ell} = \frac{\exp(e_{n,k,\ell})}{\sum_{j=1}^L \exp(e_{n,k,j})}.
\end{align}

The set $\mathbf{s}_{n,k} = [ s_{n,k,1}, \ldots, s_{n,k,L}]$ of weights defines the \textit{receptive field} of the $k^\text{th}$ spatial attention model (part detector) for image $I_n$. By definition, each receptive field is a probability mass function since $\sum_{\ell=1}^{L} s_{n,k,\ell} = 1$.  

For each image $I_n$, we generate $K$ spatial gated visual features $\{ \mathbf{x}_{n,k} \}_{k=1,K}$ using attention weighted averaging
\begin{align}
   \mathbf{x}_{n,k} = \sum_{\ell=1}^L s_{n,k,\ell} \mathbf{f}_{n,\ell}.
\label{eqn:spatial-attention}
\end{align}

\begin{figure}[t]
\begin{center}
\includegraphics[width=0.9\linewidth]{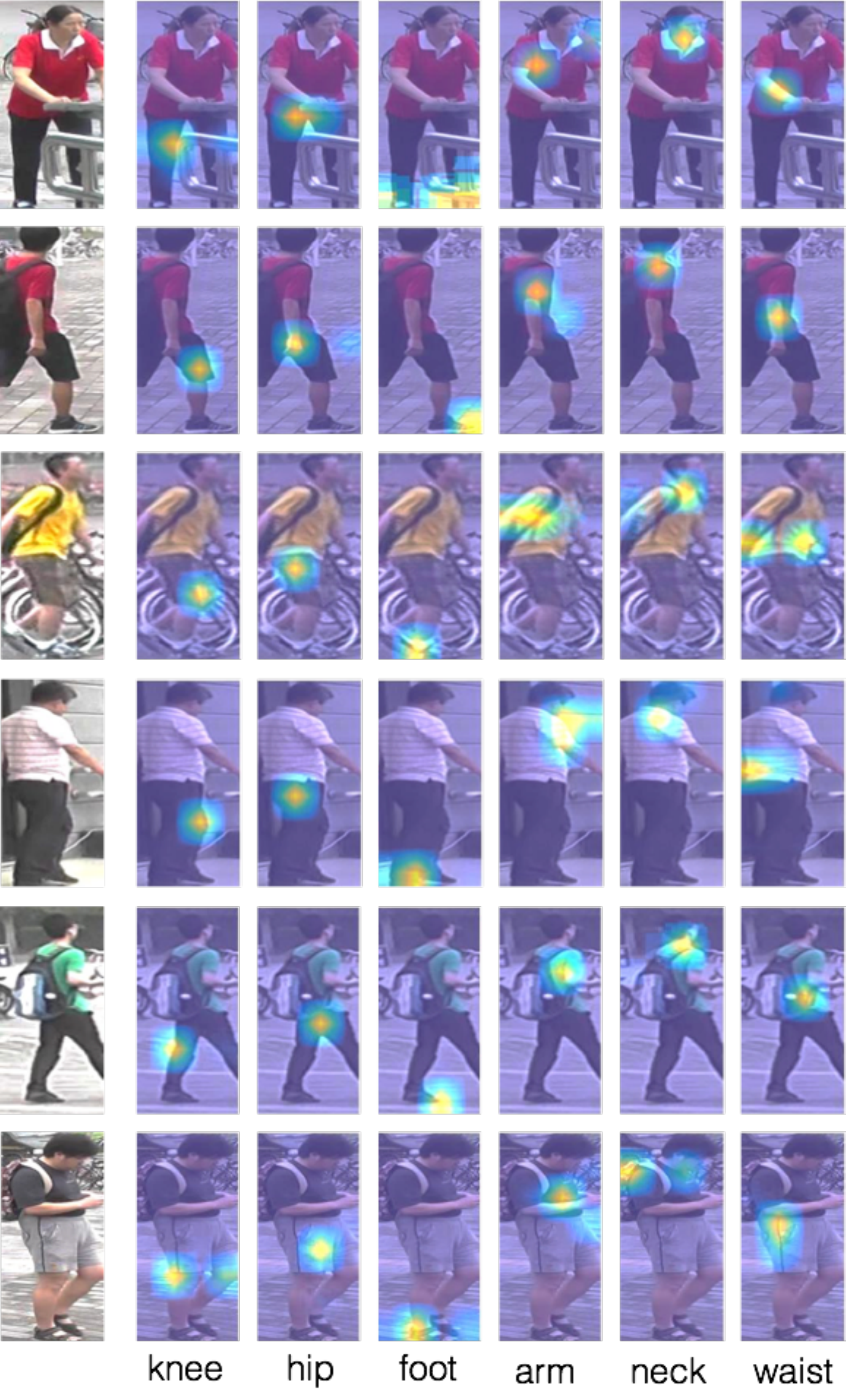} \ \\
\end{center}
\vspace{-3mm}
\caption{\textbf{Learned Spatial Attention Models}. Example images and corresponding receptive fields for our diverse spatial attention models when $K=6$. Our methodology discovers distinctive image regions which are useful for re-identification.  The attention models primarily focus on foreground regions and generally correspond to specifc body parts.  Our interpretation of each is indicated at the bottom of each column.}
\label{vis}
\end{figure}

\noindent Each gated feature represents a salient part of the input image (Fig.~\ref{vis}).  Because $\mathbf{x}_{n,k}$ is computed by pooling over the entire grid $\ell \in [1,L]$, the spatial gated feature contains no information about the image location from which it was extracted.  As a result, the spatial gated features generated for a particular attention model across multiple images are all roughly aligned---\eg extracted patches of the face all tend to have the eyes in roughly the same pixel location.


Similar to fine-grained object recognition \cite{lin2017bilinear}, we pool information across frames to created an enhanced variant 
\begin{align}
\widehat{\mathbf{x}}_{n,k} = E(\mathbf{x}_{n,k})
\label{eqn:enhancement}
\end{align} 

\noindent of each spatial gated feature. 
The enhancement function $E()$ follows the past work on second-order pooling \cite{Carreira2012}. 
See the supplementary material for further details.

\subsection{Diversity Regularization}
\label{sec:diversity}

The outlined approach for learning multiple spatial attention models can easily produce a degenerate solution.  For a given image, there is no constraint that the receptive field generated by one attention model needs to be different from the receptive field of another model. In other words, multiple attention models could easily learn to detect the same body part.  In practice, we need to ensure each of the $N$ spatial attention models focuses on different regions of the given image.


Since each receptive field $\mathbf{s}_{n,k}$ has a probabilistic interpretation, one solution is to use the Kullback-Leibler divergence to evaluate the diversity of the receptive fields for a given image. For notational convenience, we define the matrix $\mathbf{S}_n \in \mathbb{R}^{K \times L}$ as the collection of receptive fields generated for image $I_n$ by the $K$ spatial attention models
\begin{align}
    \mathbf{S}_n = [ \mathbf{s}_{n,1}, \ldots, \mathbf{s}_{n,K}].
\end{align}

\noindent Typically, the attention matrix has many values close to zero after the $\text{softmax}()$ function, and these small values drop sharply when passed though the $\log()$ operation in the Kullback-Leibler divergence. In this case, the empirical evidence suggests the training process is unstable \cite{lin2017structured}. 

To encourage the spatial attention models to focus on different salient regions, we design a penalty term which measures the overlap between different receptive fields.
Suppose $\mathbf{s}_{n,i}$ and $\mathbf{s}_{n,j}$ are two attention vectors in attention matrix $\mathbf{S}_n$.  
Employing the probability mass property of attention vectors, we use the Hellinger distance \cite{beran1977minimum} to measure the similarity of $\mathbf{s}_{n,i}$ and $\mathbf{s}_{n,j}$. The distance is defined as
\begin{align}
   H(\mathbf{s}_{n,i}, \mathbf{s}_{n,j}) = &\,\,  \frac{1}{\sqrt2} \sqrt{\sum_{\ell=1}^L (\sqrt { s_{n,i,\ell}} - \sqrt {s_{n,j,\ell}})^2}, \\
            = &\,\,  \frac{1}{\sqrt2} \Vert \sqrt {\mathbf{s}_{n,i}} - \sqrt {\mathbf{s}_{n,j}} \Vert_2.
\end{align}
Since $\sum_{\ell=1}^L s_{n,k,\ell}=1$:
\begin{align}
   H^2(\mathbf{s}_{n,i}, \mathbf{s}_{n,j}) = 1 - \sum_{\ell=1}^L (\sqrt{s_{n,i,\ell} s_{n,j,\ell}}).
\end{align}
To ensure diversity of the receptive fields, we need to maximize the distance between $\mathbf{s}_{n,i}$ and $\mathbf{s}_{n,j}$, which is equivalent to minimizing $1-H^2(\mathbf{s}_{n,i}, \mathbf{s}_{n,j})$. We introduce $\mathbf{R}_n = \sqrt{ \mathbf{S}_n }$ for notation convenience, where each element in $\mathbf{R}_n$ is the square root of the corresponding element in $\mathbf{S}_n$. Thus, the regularization term to measure the redundancy between receptive fields per image is
\begin{align}
   Q = \Vert ( \mathbf{R}_n \mathbf{R}_n^\mathsf{T}-\mathbf{I}) \Vert_F^2,
\end{align}
where $\Vert \cdot \Vert_F$ denotes the Frobenius norm of a matrix and $\mathbf{I}$ is a $K$-dimensional identity matrix. This regularization term $Q$ will be multiplied by a coefficient, and added to the original OIM loss. 

Diversity regularization was recently employed for text embedding using recurrent networks \cite{lin2017structured}.  In this case, the authors employed a variant 
\begin{align}
Q' = \Vert ( \mathbf{S}_n \mathbf{S}_n^\mathsf{T}-\mathbf{I}) \Vert_F^2
\end{align}

\noindent of our proposed regularization. Although $Q$ and $Q'$ have similar formulations, the regularization effects are very different. $Q$ is based on probability mass distributions with the constraint $\sum_{\ell=1}^L \mathbf{s}_{n,k,\ell}=1$ while $Q'$ can be formulated on any matrix. $Q'$ encourages $\mathbf{S}_n$ to be sparse -- preferring only non-zero elements along the diagonal of $\mathbf{S}_n$.  
Although $Q'$ forces the receptive fields not to overlap, it also encourages them to be concentrated to a single cell.  $Q$, on the other hand, allows large salient regions like ``upperbody'' while discouraging receptive fields from overlapping. We compare the performances of the two regularization terms $Q$ and $Q'$ in Section~\ref{sec:component-analysis}.

\subsection{Temporal Attention}
\label{sec:temporal}

Recall that each frame $I_n$ is represented by a set $\{ \widehat{\mathbf{x}}_{n,1}, \ldots, \widehat{\mathbf{x}}_{n,K} \}$ of $K$ enhanced spatial gated features, each generated by one of the $K$ spatial attention models.  We now consider how best to combine these features extracted from individual frames to produce a compact representation of the entire input video. 

All parts of an object are seldom visible in every video frame---either because of self-occlusion or from an explicit foreground occluder (Fig.~\ref{intro}).  Therefore, pooling features across time using a per-frame weight $t_n$ is not sufficiently robust, since some frames could contain valuable partial information about an individual (\eg face, presence of a bag or other accessory, etc.). 

Instead of applying the same temporal attention weight $t_n$ to all features extracted from frame $I_n$, we apply multiple temporal attention weights $ \{ t_{n,1}, \ldots, t_{n,K} \} $ to each frame---one for each spatial component.  With this approach, our temporal attention model is able to assess the importance of a frame based on the merits of the different salient regions. Temporal attention models which only operate on whole frame features could easily lose fine-grained cues in frames with moderate occlusion.
 
Similarly, basic temporal aggregation techniques (compared to temporal attention models) like average pooling or max pooling generally weaken or over emphasize the contribution of discriminative features (regardless of whether the pooling is applied per-frame, or per-region).  In our experiments, we compare our proposed per-region-per-frame temporal attention model to average and maximum pooling applied on a per-region basis, and indeed find that maximum performance is achieved with our temporal attention model. 

Similar to spatial attention, we define the temporal attention $t_{n,k}$ for the spatial component $k$ in frame $n$ to be the softmax of a linear response function
\begin{align}
    e_{n,k} = (\mathbf{w}_{t,k})^\mathsf{T} \widehat{\mathbf{x}}_{n,k} + b_{t,k},
\end{align}

\noindent where $\widehat{\mathbf{x}}_{n,k} \in \mathbb{R}^D$ is the enhanced feature of the $k^\text{th}$ spatial component in the $n^\text{th}$ frame, and $\mathbf{w}_{t,k} \in \mathbb{R}^{D}$ and $b_{t,k}$ are parameters to be learned.  
%
\begin{align}
    t_{n,k} = \frac{e_{n,k}}{\sum_{j=1}^N e_{j,k}}.
\end{align}

The temporal attentions are then used to gate the enhanced spatial features on a per component basis by weighted averaging
\begin{align}
\mathbf{x}_k &= \sum_{n=1}^N t_{n,k} \widehat{\mathbf{x}}_{n,k}.
\label{eqn:temporal-attention}
\end{align}

Combining \eqref{eqn:spatial-attention}, \eqref{eqn:enhancement} and \eqref{eqn:temporal-attention} summarizes how we apply attention on a spatial then temporal basis to extract and align portions of each raw feature $\mathbf{f}_{n,\ell}$ and then aggregate across time to produce a latent representation of each distinctive object region/part  
\begin{align}
\mathbf{x}_k &= \sum_{n=1}^N t_{n,k} E\left( \sum_{\ell=1}^L s_{n,k,\ell} \mathbf{f}_{n,\ell} \right).
\end{align}

Finally, the entire input video is represented by a feature vector $\mathbf{x} \in \mathbb{R}^{K \times D}$ generated by concatenating the temporally gated features of each spatial component
\begin{align}
    \mathbf{x} = [\mathbf{x}_1, \dots, \mathbf{x}_K].
\end{align}

\subsection{Re-Identification Loss}

In this paper, we adopt the Online Instance Matching loss function (OIM) \cite{xiao2017joint} to train the whole network. 
Typically, re-identification uses a multi-class softmax layer as the objective loss.  Often, the number of mini-batch samples is much smaller than the number of identities in the training dataset, and network parameter updates can be biased.
Instead, the OIM loss function uses a lookup table to store features of all identities appearing in the training set. 
In each forward iteration, a mini-batch sample is compared against all the identities when computing classification probabilities. 
This loss function has shown to be more effective than softmax when training re-identification networks.


\section{Experiments}


\subsection{Datasets}
We evaluate the proposed algorithm on three commonly used video-based person re-identification datasets: PRID2011 \cite{hirzer2011person}, iLIDS-VID \cite{wang2014person}, and MARS \cite{zheng2016mars}.
PRID2011 consists of person videos from two camera views, containing $385$ and $749$ identities, respectively. Only the first $200$ people appear in both cameras. The length of each image sequence varies from 5 to 675 frames. 
iLIDS-VID consists of 600 image sequences of 300 subjects. 
For each person we have two videos with the sequence length ranging from 23 to 192 frames with an average duration of 73 frames. 
The MARS dataset is the largest video-based person re-identification benchmark with 1,261 identities and around 20,000 video sequences generated by DPM detector \cite{felzenszwalb2010object} and GMMCP tracker \cite{dehghan2015gmmcp}. Each identity is captured by at least 2 cameras and has 13.2 sequences on average. There are 3,248 distractor sequences in the dataset. 

For PRID2011 and iLIDS-VID datasets, we follow the evaluation protocol from \cite{wang2014person}.
Datasets are randomly split into probe/gallery identities. 
This procedure is repeated $10$ times for computing averaged accuracies.
For the MARS dataset, we follow the original splits provided by \cite{zheng2016mars} which use the predefined 631 identities for training and the remaining identities for testing.

\subsection{Implementation details and evaluation metrics}

We divide each input video sequence into $N=6$ chunks of equal duration. 
We first pretrain the ResNet-50 model on image-based person re-identification datasets, including CUHK01 \cite{li2013locally}, CUHK03 \cite{li2014deepreid}, 3DPeS \cite{baltieri20113dpes}, VIPeR \cite{gray2007evaluating}, DukeMTMC-reID \cite{zheng2017unlabeled} and CUHK-SYSU \cite{xiao2017joint}.
and then fine-tune it on PRID2011, iLIDS-VID and MARS training sets. 
Once finished, we fix the CNN model and train the set of multiple spatial attention models with average temporal pooling and OIM loss function. 
Finally, the whole network, except the CNN model, is trained jointly. 
The input image is resized to $256 \times 128$. 
The network is updated using batched Stochastic Gradient Descent with an initial learning rate set to $0.1$ and then dropped to $0.01$. The aggregated feature vector after the last FC layer is embeded into $128$-dimensions and L2-normalized to represent each video sequence. During the training stage, we utilize the Restricted Random Sampling to select training samples. For each video, we extract its L2-normalized feature and sent it to the OIM loss function to supervise the training process.
During testing, we use the first image from each of $N$ segments as a testing sample and its L2-normalized features are utilized to compute the similarity of the spatiotemporal gated features generated for the pair of videos being assessed. 


Re-identification performance is reported using the rank-1 accuracy.
On the MARS dataset we also evaluate the mean average precision (mAP) \cite{zheng2016mars}. 
Since mAP takes recall into consideration, it is more suitable for the MARS dataset which has multiple videos per identity.

\subsection{Component Analysis of the Proposed Model}
\label{sec:component-analysis}
\begin{table}
\center
\begin{adjustbox}{max width=0.48\textwidth}
\begin{tabular}{lccc} 
\textsc{Method} & PRID2011 & iLIDS-VID & MARS \\
\midrule
\textbf{Baseline}            	&82.7    &61.2 		&73.4 (58.1) \\ 
\textbf{SpaAtn}              	&84.2    &64.9    	&74.5 (59.3) \\ 
\textbf{SpaAtn+Q$'$}        		&86.5    &64.5    	&74.0 (58.2) \\ 
\textbf{SpaAtn+Q}        		&86.7    &68.6    	&77.0 (60.9) \\ 
\textbf{SpaAtn+Q+MaxPool} 	&86.9    &68.2      &76.8 (60.5) \\ 
\textbf{SpaAtn+Q+TemAtn} 	&88.4    &69.7      &77.1 (61.2) \\ 
\textbf{SpaAtn+Q+TemAtn+Ind} 	&\textbf{93.2}    &\textbf{80.2}      &\textbf{82.3} (\textbf{65.8}) \\ 
\bottomrule 
\end{tabular}
\end{adjustbox}
\vspace{5pt}
\caption{Component analysis of the proposed method: rank-1 accuracies are reported. 
For MARS we provide mAP in brackets.
\textbf{SpaAtn} is the multi-region spatial attention, \textbf{Q$'$} and \textbf{Q} are two regularization terms, \textbf{MaxPool} and \textbf{TemAtn} are max temporal pooling and the proposed temporal attention respectively. \textbf{Ind} represents fine-tuning the whole network to each dataset independently.}
\label{tab:component}
\end{table}

We investigate the effect of each component of our model by conducting several analytic experiments. 
In Tab.~\ref{tab:component}, we list the results of each component in the proposed network. 
\textbf{Baseline} corresponds to ResNet-50 trained with OIM loss on image-based person re-id datasets and then jointly fine-tuned on video datasets: PRID2011, iLIDS-VID, and MARS.
\textbf{SpaAtn} consists of the subnetwork of ResNet-50 (from $res2x$ to $res5x$) and multiple spatial attention models. 
All spatial gated features generated by the same attention model are grouped together and averaged over all frames. 
For each video sequence, there will be $K$ averaged feature vectors. 
We concatenate the $K$ features and then send them to the last FC layer and OIM loss function to train the neural network. 
Compared with \textbf{Baseline}, \textbf{SpaAtn} improves the rank-1 accuracy by $1.5 \%$, $3.7 \%$, and $1.1 \%$ on PRID2011, iLIDS-VID and MARS, respectively. 
This shows that multiple spatial attention models are effective at finding persistent discriminative image regions which are useful for boosting re-identification performance.

\textbf{SpaAtn+Q'} has the same network architecture as \textbf{SpaAtn} but with the text embedding diversity regularization term $Q'$ \cite{lin2017structured}. 
\textbf{SpaAtn+Q} uses our proposed diversity regularization term $Q$ based on Hellinger distance. From the results, we can see that our proposed Hellinger regularization improves accuracy. 
We believe the improvement comes from being able to learn
multiple attention models with sufficiently large (but minimally overlapping) 
receptive fields (see Fig.\ref{vis} for sample receptive fields generated for the learned attention models using \textbf{SpaAtn+Q}).
\textbf{SpaAtn+Q} and \textbf{SpaAtn+Q+MaxPool} are strategies for average temporal pooling and maximum temporal pooling, respectively.
\textbf{SpaAtn+Q+TemAtn} applies multiple temporal attentions to each frame---one for each diverse spatial attention model. The assigned temporal attention weights reflect the pertinence of each spatially attended region (\eg is the part fully visible and easy to detect?). 
We finally fine-tune the whole network, including the CNN model, to each video dataset independently.
\textbf{SpaAtn+Q+TemAtn+Ind} is the final result of our proposed framework.\\

\noindent
\textbf{Different number of spatial attention models:}
\begin{table}
\center
\begin{adjustbox}{max width=0.48\textwidth}
\begin{tabular}{lccc} 
\textsc{$K$} & PRID2011 & iLIDS-VID & MARS \\
\midrule
1   &86.2     &64.7    &76.0  \\ 
2   &83.4     &64.6    &75.7  \\ 
4   &86.9     &64.6    &\textbf{77.2} \\ 
6   &\textbf{88.4}  &\textbf{69.7}     &77.1 \\ 
8   &88.0     &66.9   &76.7 \\ 
\bottomrule 
\end{tabular}
\end{adjustbox}
\vspace{5pt}
\caption{The rank-1 accuracy using different number $K$ of diverse spatial attention models.}
\label{tab:number}
\end{table}
We also carry out experiments to investigate the effect of varying the number $K$ of spatial attention models (Tab.~\ref{tab:number}).
When $K=1$, the framework is limited to a single spatial attention model, which tends to cover the whole body.  As $K$ is increased, the network is able to discover a larger set of body parts, and since the receptive fields are regularized to have minimal overlap, the reception fields tend to shrink as $K$ gets bigger.  Interestingly, there is a general drop in perform when $K$ is increased from $1$ to $2$. This implies treating a person as a single region instead of two distinct body parts is better. However, when a sufficiently large $K=6$ number of spatial models is used, the network achieves maximum performance. 

Example learned spatial attention models and corresponding receptive fields are shown in Fig.~\ref{vis}.  The receptive fields generally correspond to specific body parts and have varying sizes dependent on the discovered concept.  In constrast, the receptive fields generated by \cite{liu2017hydraplus} tend to include background clutter and exhibit substantial overlap between different attention models.  Our receptive fields, on the other hand, have minimal overlap and focus primarily on the foreground regions.

\subsection{Comparison with the State-of-the-art Methods}

\begin{table}
\center
\begin{adjustbox}{max width=0.48\textwidth}
\begin{tabular}{lrrr} 
\textsc{Method} & PRID2011 & iLIDS-VID & MARS \\
\midrule
STA \cite{liu2015spatio}                  &64.1     &44.3   & -  \\ 
DVDL \cite{karanam2015person}             &40.6     &25.9   & -  \\ 
TDL \cite{you2016top}                     &56.7     &56.3   & -  \\ 
SI2DL \cite{zhu2016video}                 &76.7     &48.7   & - \\ 
mvRMLLC+Alignment \cite{chen2016person}   &66.8     &69.1   & - \\ 
AMOC+EpicFlow \cite{liu2017video}         &82.0     &65.5   & - \\ 
RNN \cite{mclaughlin2016recurrent}        &70.0     &58.0   & - \\ 
IDE \cite{zheng2016person} + XQDA \cite{liao2015person}  &- &- &65.3 (47.6) \\ 
GEI+Kissme \cite{zheng2016mars}           &19.0     &10.3   &1.2 (0.4) \\
end AMOC+EpicFlow \cite{liu2017video}     &83.7 &68.7  &68.3 (52.9) \\
Mars \cite{zheng2016mars}       &77.3  &53.0  &68.3 (49.3) \\
SeeForest \cite{zhousee}        &79.4  &55.2  &\textbf{\textcolor{blue}{70.6}} (\textbf{\textcolor{blue}{50.7}}) \\ 
QAN \cite{liu2017quality}          &90.3  &68.0  & -   \\
PAM-LOMO+KISSME \cite{khan2017multi}            &\textbf{\textcolor{blue}{92.5}}  &\textbf{\textcolor{blue}{79.5}}  & -   \\
\cmidrule{2-4}
\textbf{Ours}                             &\textbf{\textcolor{red}{93.2}} &\textbf{\textcolor{red}{80.2}} &\textbf{\textcolor{red}{82.3}} (\textbf{\textcolor{red}{65.8}}) \\

\bottomrule 
\end{tabular}
\end{adjustbox}
\vspace{5pt}
\caption{Comparisons of our proposed approach to the state-of-the-art on PRID2011, iLIDS-VID, and MARS datasets. The rank-1 accuracies are reported and for MARS we provide mAP in brackets.
The best and second best results are marked by \textcolor{red}{red} and \textcolor{blue}{blue} colors, respectively.}
\label{tab:statemethods}
\end{table}
Table~\ref{tab:statemethods} reports the performance of our approach
with other state-of-the-art techniques. 
On each dataset, our method attains the highest performance. 
We achieve maximum improvement on MARS dataset, where we improve the state-of-the-art by 11.7\%.
The previous best reported results are from PAM-LOMO+KISSME \cite{khan2017multi} (which learns signature representation to cater for high variance in a person's appearance) and from SeeForest \cite{zhousee} (which combines six spatial RNNs and temporal attention followed by a temporal RNN to encode the input video). In contrast, our network architecture is intuitive and straightforward to train. 
MARS is the most challenging data (it contains distractor sequences and has a substantially larger gallery set) and our methodology achieves a significant increase in mAP accuracy.  This result suggests our spatiotemporal model is very effective for video-based person re-identification in challenging scenarios.


\section{Summary}

A key challenge for successful video-based person re-identification is developing a latent feature representation of each video as a basis for making comparisons.  In this work, we propose a new spatiotemporal attention mechanism to achieve better video representations.  Instead of extracting a single feature vector per frame, we employ a diverse set of spatial attention models to consistently extract similar local patches across multiple images (Fig.~\ref{vis}).  This approach automatically solves two common problems in video re-identification: aligning corresponding image patches across frames (because of changes in body pose, orientation relative to the camera, etc.) and determining whether a particular part of the body is occluded or not.  

To avoid learning redundant spatial attention models, we employ a diversity regularization term based on Hellinger distance.  This encourages the network to discover a set of spatial attention models that have minimal overlap between receptive fields generated for each image.  Although diversity regularization is not a new topic, we are the first to learn a diverse set of spatial attention models for video sequences, and illustrate the importance of Hellinger distance for this task (our experiments illustrate how a diversity regularization term used in text embedding is less effective for images).  

Finally, temporal attention is used to aggregate features across frames on a per-spatial attention model basis---\eg all features from the facial region are combined.  This allows the network to represent each discovered body part based on the most pertinent image regions within the video.  We evaluated our proposed approach on three datasets and performed a series of experiments to analyze the effect of each component. Our method outperforms the state-of-the-art approaches by large margins which demonstrates its effectiveness in video-based person re-identification.

{\small
\bibliographystyle{ieee}
\bibliography{egbib}
}

\end{document}


\title{\textit{Supplementary Material} \\ Diversity Regularized Spatiotemporal Attention\\for Video-based Person Re-identification}

\maketitle
\thispagestyle{empty}

\setcounter{section}{3}
\setcounter{subsection}{2}

\subsubsection{Feature Enhancement}
\label{sec:enhancement}
After running multi-region spatial attention, each frame is represented by $K$ spatially gated features. 
We then pool these features across time to enhance their representation.
Recall that $\mathbf{x}_{n,k} \in \mathbb{R}^D$ is the feature vector of the $k^\text{th}$ spatial component from the $n^\text{th}$ frame.  
The matrix $\mathbf{X}_k = [ \mathbf{x}_{1,k}, \dots, \mathbf{x}_{N,k} ], \mathbf{X}_k \in \mathbb{R}^{D \times N}$ is defined as the set of spatial gated visual features generated by the $k^\text{th}$ attention model over all frames.  
Because frames with similar spatial gated features tend to represent multiple observations of the same patch of pixel values, we generate a robust feature representation of each component at each frame by pooling information from the same component at other frames.  
For each feature extracted at frame $n$, we pool information from auxiliary frames by assigning a weight to each of them, 
which depends on the similarity of the spatial gated features as well as the temporal distance between the two frames.

We define a per-component ``feature similarity'' matrix $\mathbf{\Phi}_k \in \mathbb{R}^{N \times N}$ as the inner product of feature vectors
\begin{align}
\mathbf{\Phi}_k = (\mathbf{X}_k)^\mathsf{T} \mathbf{X}_k.
\end{align}
Additionally, we define a ``temporal similarity'' matrix $\Psi \in \mathbb{R}^{N \times N}$
\begin{align}
   \mathbf{\Psi} = \mathbf{W} \exp \left( -\frac{|i-j|}{\sigma} \right) + \mathbf{b},
\end{align}
where
$|i-j|$ is a matrix that represents relative temporal distance between any frame $i$ and frame $j$, and $\mathbf{W} \in \mathbb{R}^{N \times N}$ and $\mathbf{b} \in \mathbb{R}^N$ encodes the positional information.




The overall similarity scores across all frames in the video sequence of the $k^\text{th}$ spatial component are represented as follows. 
\begin{align}
   \mathbf{C}_k = \text{softmax}'(\mathbf{\Phi}_k + \mathbf{\Psi} ),
\label{sim_app}
\end{align}
where $\mathbf{\Phi}_k$ and $\mathbf{\Psi}$ formulate the appearance and location similarity respectively. $\text{Softmax}'$ is the Softmax function over each row of $(\mathbf{\Phi}_k + \mathbf{\Psi} )$. Elements in the $i^\text{th}$ row of $\mathbf{C}_k$ describe the contribution probabilities of all frames in the input sequence to frame $i$.
The enhanced feature representation is written as a residual connection,
\begin{align}
    \mathbf{\widehat{X}}_k = FCN(\mathbf{X}_k \mathbf{C}_k) + \mathbf{X}_k,
\end{align}
where $FCN$ are a linear transformations. $\widehat{\mathbf{x}}_{n,k}  \in \mathbb{R}^D$ is the $n^\text{th}$ element of $\mathbf{\widehat{X}}_k$ which corresponds to Eq.~(4) in the paper (${\widehat{\mathbf{x}}_{n,k} = E(\mathbf{x}_{n,k})}$).



\setcounter{section}{4}
\setcounter{subsection}{4}

\subsection{More Experimental Results}
\begin{figure*}[t]
\begin{center}
\includegraphics[width=0.9\linewidth]{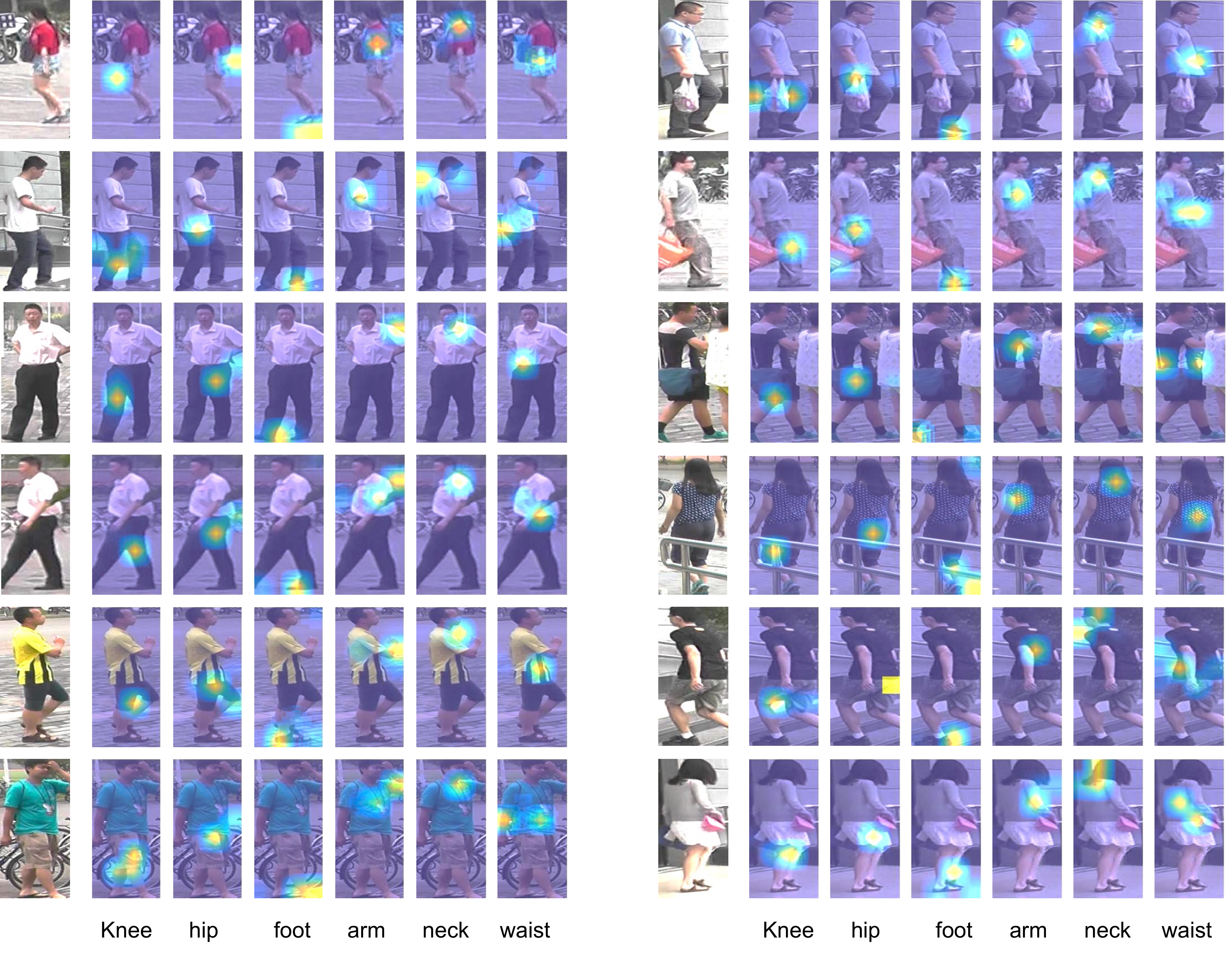} \ \\
\end{center}
\vspace{-3mm}
\caption{Example images and corresponding receptive fields for our diverse spatial attention models when $K=6$. Our methodology discovers distinctive image regions which are useful for re-identification.  The attention models primarily focus on foreground regions and generally correspond to specifc body parts.  Our interpretation of each is indicated at the bottom of each column.}
\label{vis}
\end{figure*}

\begin{figure*}[t]
\begin{center}
\includegraphics[width=0.7\linewidth]{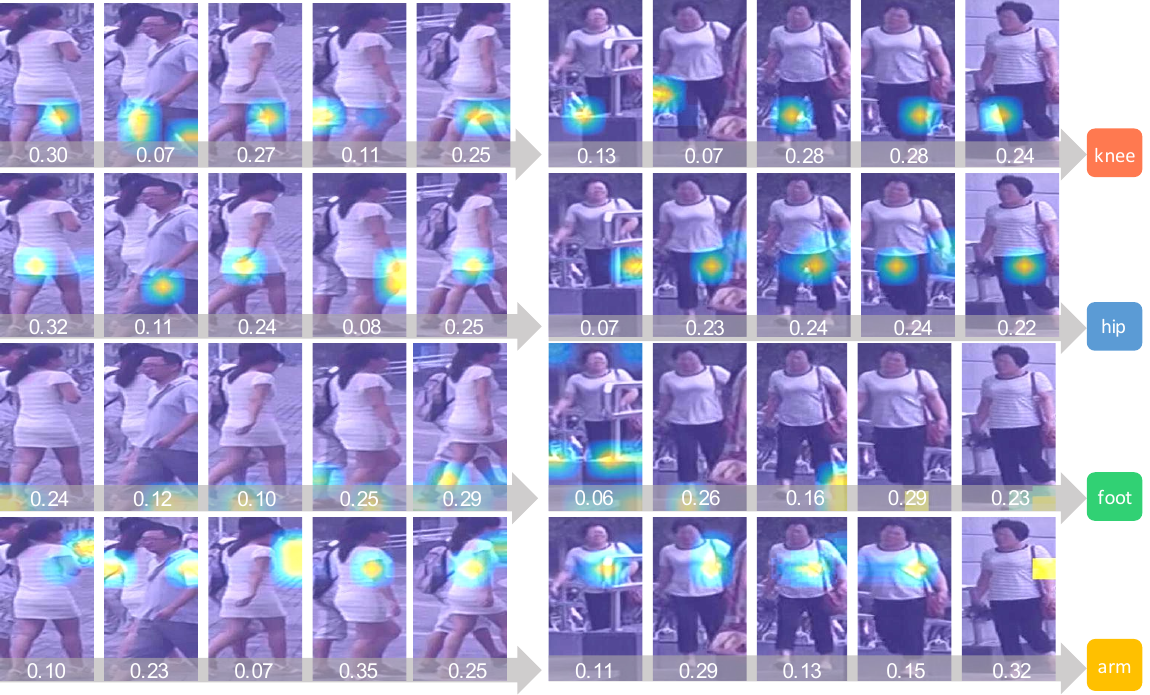} \ \\
\end{center}
\vspace{-5pt}
\caption{Visualization results. The numbers under images indicate the temporal attention weights assigned to each frame. Our temporal attentions assign low weights to occluded or background parts and high weights to correctly detected parts.}
\label{rebuttal}
\end{figure*}

Fig.\ref{vis} illustrates more examples of corresponding receptive fields generated by multiple spatial attention models.
Fig.\ref{rebuttal} is the visualization results of temporal attention. The temporal attention model learns the importance of each frame based on the merits of different salient regions. Our temporal attentions assign low weights to occluded or background parts and high weights to correctly detected parts which confirms the effectiveness of the model.

